# MedArena: Comparing LLMs for Medicine-in-the-Wild Clinician Preferences

Eric Wu[1], Kevin Wu[2], Jason Hom[3], Paul H. Yi[4], Angela Zhang[5], Alejandro Lozano[2], Jeff Nirschl[6], Jeff Tangney[7], Kevin Byram[8], Braydon Dymm[9], Narender Annapureddy[8], Eric Topol[10], David Ouyang[11], James Zou[1,2]

[1]Department of Electrical Engineering, Stanford University, Stanford, CA, USA
[2]Department of Biomedical Data Science, Stanford University, Stanford, CA, USA
[3]Division of Hospital Medicine, Department of Medicine, Stanford School of Medicine, Stanford, CA, USA
[4]Department of Radiology, St. Jude Children's Research Hospital, Memphis, TN, USA
[5]University of California, San Francisco, San Francisco, CA, USA
[6]Department of Pathology and Laboratory Medicine, University of Wisconsin School of Medicine and Public Health, Madison, WI, USA
[7]Doximity, San Francisco, CA, USA
[8]Department of Medicine, Division of Rheumatology and Immunology, Vanderbilt University Medical Center, Nashville, TN, USA
[9]Department of Neurology, Charleston Area Medical Center, Charleston, WV, USA
[10]Department of Translational Medicine, Scripps Research Translational Institute, La Jolla, CA, USA
[11]Kaiser Permanente Division of Research, Pleasanton, CA, USA

## Abstract

Large language models (LLMs) are increasingly central to clinician workflows, spanning clinical decision support, medical education, and patient communication. However, current evaluation methods for medical LLMs rely heavily on static, templated benchmarks that fail to capture the complexity and dynamics of real-world clinical practice, creating a dissonance between benchmark performance and clinical utility. To address these limitations, we present *MedArena*, an interactive evaluation platform that enables clinicians to directly test and compare leading LLMs using their own medical queries. Given a clinician-provided query, MedArena presents responses from two randomly selected models and asks the user to select the preferred response. Out of 1571 preferences collected across 12 LLMs up to November 1, 2025, Gemini 2.0 Flash Thinking, Gemini 2.5 Pro, and GPT-4o were the top three models by Bradley-Terry rating. Only one-third of clinician-submitted questions resembled factual recall tasks (e.g., MedQA), whereas the majority addressed topics such as treatment selection, clinical documentation, or patient communication, with ~20% involving multi-turn conversations. Additionally, clinicians cited *depth and detail* and *clarity of presentation* more often than raw factual accuracy when explaining their preferences, highlighting the importance of readability and clinical nuance. We also confirm that the model rankings remain stable even after controlling for style-related factors like response length and formatting. By grounding evaluation in real-world clinical questions and preferences, MedArena offers a scalable platform for measuring and improving the utility and efficacy of medical LLMs.

**Introduction**

The use of large language models (LLMs) in the medical domain holds transformative potential, promising advances across clinical decision support, medical education, and patient communication. This increasing relevance is highlighted by [recent reports](recent reports) indicating that up to two-thirds of American physicians now utilize AI tools in their practice. Furthermore, recent US Food and Drug Administration (FDA) guidance suggests LLMs are not considered software-as-a-medical device and may be regulated with a lighter touch.

Realizing the potential of LLMs to assist clinicians safely and effectively hinges on the development of rigorous and clinically relevant evaluation methodologies. Currently, the predominant approaches for assessing the medical capabilities of LLMs, such as benchmark datasets derived from MMLU (Massive Multitask Language Understanding) and MedQA (Medical Question Answering), primarily utilize *static*, *multiple-choice question* (MCQ) formats. While valuable for gauging foundational knowledge, the static, closed-ended, and templated design of current benchmarks suffers from significant challenges that limit their applicability to real-world clinical contexts.

In real-world clinical contexts, clinicians encounter scenarios that are often open-ended, dynamic, multi-step, and span a wide range of clinical tasks. For example, clinicians rarely have all the diagnostic information needed to optimally identify the next diagnostic or management step, requiring tools that can assist in scenarios involving substantial uncertainty. Additionally, clinical scenarios are often dynamic and multi-step as clinical scenarios evolve in response to continuous new information (from the history, serial physical exams, laboratory tests, imaging or other sources) and clinical decisions being made, requiring tools that are capable of multi-turn conversations. Finally, the responsibilities of a clinician extend beyond selecting the next best step to include interpreting multimodal clinical results, summarizing clinical courses, communicating with patients, translating clinical documents, and applying the most recent clinical guidelines and drug approvals. Existing multiple-choice Q&A datasets are inadequate for evaluating the complex, open-ended tasks that clinicians require LLMs to perform in practice, providing an incomplete picture of an LLM's true clinical utility.

This highlights a critical need for evaluation frameworks that move beyond these current limitations. Specifically, medical LLM evaluations must become more *dynamic*, capable of reflecting the most current medical questions in the wild and adapting to the iterative nature of clinical questioning, and more *holistic*, assessing the entire response quality, including reasoning, multi-turn conversations, and clinical appropriateness in the context of a specific and unique case, rather than merely scoring factual accuracy on a fixed answer set.

To this end, we introduce [MedArena.ai](MedArena.ai), the first LLM evaluation platform specifically designed for clinical medicine. MedArena captures clinician preferences through head-to-head model comparisons on real-world, clinician-authored queries, enabling fine-grained analysis of model performance across diverse clinical tasks, interaction styles, and reasoning depths that are difficult or impossible to assess with traditional static benchmarks.

**Related works**

Current evaluation frameworks for medical LLMs fall into three general categories: multi-choice question (MCQ) evaluations, open-ended diagnostic and reasoning benchmarks, and task-based performance assessments.

LLMs are most commonly evaluated on static MCQ datasets [1], given the simplicity of evaluating models on multiple-choice options and the ability to compare models independently. Popular benchmark datasets such as MedQA, MedMCQA, MMLU, MedBullets, and MedXpertQA are derived from medical licensing and board exams and generally assess factual recall and basic reasoning capabilities. However, because they are static by design, they fail to adapt to new clinical guidelines or standards of care. Additionally, the constrained multiple-choice format is unrealistic in real-world clinical care settings and may mask correct guesses with spurious or even inaccurate reasoning [2].

Several recent benchmarks have been introduced to remedy these limitations. MultiMedQA [3], for instance, contains open-ended questions but relies on using LLMs as judges of correctness, a setup that can introduce biases and errors. MedCaseReasoning is an open-ended diagnostic reasoning dataset sourced from medical case reports, but similarly relies on LLM-as-a-judge. While these datasets more closely imitate real-world clinical scenarios, they are also limited to diagnostic tasks and may not reflect the broader spectrum of clinical work, such as treatment planning, patient communication, discharge documentation, or longitudinal care.

To this end, MedHelm [4] and MedS-Bench [5] are two examples of task-based benchmarks of medical LLMs and include evaluations on tasks such as case summarization, note generation, patient letter writing, and treatment recommendations. While these datasets reflect a wider spectrum of real-world clinical tasks, they remain static and cannot be updated to reflect changes in medical practice or use cases. Additionally, they cannot capture clinician-specific preferences, such aswriting style and tone.

MedArena is inspired by Chatbot Arena [6], an open-source crowdsourced platform for evaluating LLMs in head-to-head comparisons from user-provided input queries. One key distinction of MedArena is that our framework is offered exclusively to licensed clinicians, whereas Chatbot Arena does not require any user authentication. As such, MedArena reflects the preferences of medical experts and healthcare providers, whereas the Chatbot Arena leaderboard reflects the preferences of lay users. Additionally, MedArena only surfaces top commercially available LLMs, whereas Chatbot Arena allows model developers to upload anonymous preview or beta models. Indeed, a prominent criticism of Chatbot Arena is the ability of model developers to manipulate their rankings by submitting private models that have been trained specifically to perform well on Chatbot Arena [7]. MedArena is also the first platform to source head-to-head clinical preferences on open-ended questions generated by clinicians themselves.

**Methods**

*Authentication*

MedArena is open to clinicians only. We allow clinicians to authenticate their credentials through two methods. First, we partnered with Doximity, the largest networking service for medical professionals, which has over 80% of US doctors signed up on the platform. Users are allowed to sign in with their Doximity account, which is authenticated by Doximity to be linked to a licensed US clinician. Alternatively, users can provide their National Provider Identifier (NPI) number, which is verified via API at sign-up. The NPI is assigned to licensed healthcare providers in the United States. For clinicians outside of the US, they are also required to provide their credentials (e.g., national ID, institutional email, etc.), although the information is not verified due to non-standard international credentialing. Users are asked not to submit protected health information (PHI) as publicly available LLM APIs are not HIPAA-compliant.

*User interface*

For an input query, the user is presented with responses from two randomly chosen LLMs (models A and B) and asked to specify which model they prefer (Figure 1). The model responses are streamed simultaneously, and four preference options appear once both models finish their generations: Prefer model A, prefer model B, tie, or neither. In addition to these four choices, users are also provided an optional text field to specify their reason for providing their preference. The user can also continue the conversation across multiple turns before submitting a preference, in which case their preference reflects the entire multi-turn interaction rather than a single response. Once the user submits the preference, they are given the option to submit another question or to regenerate responses to the same question from two newly randomly selected models. The platform aggregates preferences from all users and presents a [leaderboard](leaderboard) (Figure 2), ranking different LLMs against each other. To help clinicians understand their own preferences, we also provide personal rankings based on a user's individual data, given a minimum number of preferences (N=20).

*Model Selection and Evaluation*

From the user preference data, we aggregate and compute rankings for all models. In total, twelve commercially available LLMs are included in MedArena (Figure 2). These models were chosen for inclusion based on overall popularity and frequency of real-world clinician usage. They include three Google Gemini [8] models, four OpenAI [9] models, three Llama [10] models (including one Perplexity model), and one Anthropic model. We were unable to include medical-specific LLMs like OpenEvidence or DoxGPT because they lack API integration. Five models additionally support images as input, and three models include retrieval augmented generation (RAG) capabilities (i.e., the ability to cite web sources in their responses).

*Categories for query use cases, subspecialty, and preference reasons*

To better understand the types of queries that physicians are asking LLMs, we categorized the clinician queries into one of six use cases:

1. Medical Knowledge and Evidence: Questions recalling biomedical facts or literature-based evidence.
2. Treatment and Guidelines: Queries about medication use, dosing, clinical protocols, and current practice recommendations.
3. Clinical Cases and Diagnosis: Case-based reasoning questions involving differential diagnoses and diagnostic decision-making.
4. Patient Communication and Education: Requests for help with explaining conditions, counseling patients, or answering patient questions in lay terms.
5. Clinical Documentation and Practical Information: Questions related to note taking, billing codes, or medical procedure codes.
6. Miscellaneous

Similarly, we categorize each query according to the most closely related of six medical subspecialties:

1. General Internal Medicine
2. Imaging-based Medicine (radiology and pathology-based queries, including both image and text-based queries)
3. Cardiology
4. Neurology and Neuropsychiatric Disorders
5. Infectious Diseases
6. Other subspecialties

Finally, on preferences that contained user-specified reasons, we also cluster them into one of five categories:

1. Accuracy and Clinical Validity: The preferred response is more accurate and clinically valid (e.g., "Model A is right")
2. Depth and Detail: The preferred response provides more depth and detail.
3. Presentation and Clarity: The preferred response is better presented and easier to understand.
4. Use of References and Up-to-Date Guidelines: The preferred response uses references and up-to-date guidelines.
5. Miscellaneous: The preferred response is not categorized in the above categories.

For each of the three categorizations, we prompt *gpt-4o-mini* to assign the query to one of the categories. The prompts are provided in Supp. Figure 6.

*Model Ranking and Statistical Analysis*

To compute quantitative rankings over the model preferences, we introduce two methods: first, we use the Elo rating system [11], which is a popular non-parametric approach that assigns ratings sequentially based on the outcomes of head-to-head matchups. For each matchup, the winning model's Elo rating is increased (and the losing model's rating is decreased) inversely proportional to the difference in rating prior to the matchup. We assign a base rating of 1000 and choose a K factor of 4, identical to the hyperparameters used in Chatbot Arena [6]. An alternative to Elo is the Bradley-Terry (BT) model [12]. Instead of updating ratings sequentially based on match outcomes, BT treats the entire set of pairwise comparisons as a probabilistic model, where each model is assigned a latent score and the probability of one model beating another is a logistic function of the score difference. The BT model is fit using maximum likelihood estimation over all observed preferences, enabling a global optimization that can handle incomplete or imbalanced comparison data more robustly than Elo.

We additionally compute 95% bootstrapped confidence intervals on the ratings, as well as the win rate. Based on the win rate, we report the p-value versus the next highest model (i.e., the proportion of times via sampling that the model wins against the next lower-ranked model with N=1000).

*Confounder analysis*

To disentangle substantive content differences from spurious presentation variations in model outputs, we perform confounder analysis on time- and style-related covariates to determine the extent to which user preferences are influenced by superficial attributes rather than core informational quality [6].

First, we extend the BT model by explicitly modeling several style-related variables:

1. Response length - number of characters
2. Headers - number of occurrences of '#'
3. Lists - number of occurrences of '- ' or '* '
4. Bolded text - number of occurrences of '**'
5. Citations - whether the response contains citations (0 or 1)

Pairwise preferences between models are treated as probabilistic outcomes modeled by the logistic function of two components: a latent skill rating difference and a weighted linear combination of stylistic feature differences. Each matchup includes derived features such as character count, header usage, list usage, and bold text usage, computed separately for both model responses and normalized by their sum. These deltas are standardized and used as covariates. The model jointly optimizes the latent ratings and feature coefficients via L-BFGS to minimize regularized negative log-likelihood. Bootstrap resampling (N=100) is used to compute 95% confidence intervals for both model ratings and feature weights. Feature importance is assessed via median coefficient magnitude and two-sided bootstrap p-values based on sign consistency. To assess whether response length systematically influenced user preferences, we conducted a nonparametric Mann–Whitney U test comparing the character lengths of preferred responses versus non-preferred responses.

**Results**

*MedArena users reflect an experienced, diverse clinician base*

As of November 1, 2025, 357 credentialed users have signed up and submitted at least one preference to MedArena (Supp. Figure 1). Of the 106 unique subspecialties represented, the top three most common are internal medicine (13%), family medicine (7%), and general practice physician (5%). Among clinicians who specified their credentials, over 84% are MD/DOs, while the rest are PAs, NPs, and other professions. 62% of clinicians have five or more years of licensed experience, with 19% having more than 20 years of experience.

*Google Gemini and OpenAI models top leaderboard rankings*

Since our general release in early March to November 1, 2025, MedArena.ai has collected 1571 preferences across 12 LLMs. The top three models are Gemini 2.0 Flash Thinking, Gemini 2.5 Pro, and GPT-4o (Table 1). Interestingly, some non-reasoning models like GPT-4o outperformed their reasoning counterparts like gpt-o1. Gemini 2.0 Flash Thinking has the top BT rating of 1110. BT and Elo ratings are highly correlated (Pearson R=0.95). The lowest-rated model is Anthropic's Claude 3.5 Sonnet, which has an Elo rating of 848.

**Style differences in model responses do not significantly affect model rankings**

After controlling for style variables, we found that the Pearson R correlation between the BT rating with and without style control is very high (0.96). For instance, we observe that Gemini 2.0 Flash Thinking drops from 1st to 3rd, and Gemini 2.5 Pro goes from 2nd to 1st (Supp. Table 2). The frequency of bold text and lists in model responses was found to be statistically significant (p-value < 0.05) covariates in the BT model, but response length, headers, and presence of citations were not (Supp. Table 1). We find that response length has a statistically significant effect on user preferences: preferred responses had a median character length of 4386 compared to 3804 for non-preferred ones, with a Mann–Whitney U test yielding *p* < 0.0001. However, when integrated into the full contextual BT model, length was not a significant predictor of final model rankings—its coefficient was near zero with wide bootstrap confidence intervals overlapping zero—indicating that while users may have superficial preferences for longer outputs, length does not meaningfully contribute to the underlying perception of model quality when controlling for other factors.

*Clinician queries span a wide range of use cases*

Our results also highlight the mismatch between existing benchmark tasks and the actual types of questions clinicians ask (Figure 3). A third of real-world queries fell into the traditional use case of medical knowledge and evidence, the focus of most current evaluations (e.g., MedQA, MMLU). Examples of these types of queries include "*Will air trapping resolve on ct with prone imaging?*" and *"Is intracranial stenting routinely performed for stroke treatment?"* (Supp. Figure 3). Treatment and Guidelines also account for about one-third of queries. These queries tend to be more detailed and patient-specific (e.g., "*Treatment strategy for HLH in a 35yo with negative extensive workup including malignancy, infection, autoimmune, and genetic testing?*"). Clinical Cases and Diagnoses make up about a fifth of queries, and often involve short clinical vignettes: "*An otherwise healthy 50 year old patient presents with a first time seizure. Labs are unremarkable. What is the next appropriate step in the diagnostic workup?*" Patient

Communication and Education queries involve queries for improving patient care like: *"A medical resident asks for help explaining to patients the difference between tension headaches, migraines, and cluster headaches. Provide a simple framework that contrasts key features, pathophysiology, and management approaches."* Finally, queries in the Clinical Documentation and Practical Information category may ask for specific documentation templates, e.g., *"create a structured template for Patient for infertility workup"*. More examples are included in Figure 5. Our analysis reveals that the majority of clinician queries centered on practical, context-rich areas like treatment decision-making, patient communication, and documentation -- domains poorly captured by static MCQs. Additionally, about 20% of conversations were multi-turn, meaning that they involved the clinician responding to the model output before providing a preference. Multi-turn conversations are also not well-captured in current evaluation benchmarks. This divergence underscores the need for evaluation systems grounded in real clinical workflows.

*Clinicians emphasize detail and accuracy in model response preferences*

Of the 1571 preferences submitted through MedArena, 153 (9.7%) of them included clinician-provided free-text reasons for their model choice (Figure 4). The most common provided reason for a preference (~⅓) is the Depth and Detail of the model response (e.g., *"Model A/B breaks down the causes better"* or *"Model A/B has more detailed information"*). Of note, our analysis of style-related differences revealed that response length alone cannot significantly explain model preference, so the detail that clinicians are referring to are likely more nuanced than simply the presence of more text. The second most provide reason is Accuracy and Clinical Validity (about one-fifth). Here, clinicians provided the most in-depth reasons, including query-specific analysis on comparing model responses. One example mentions an incorrect hallucination in one of the model responses: *"Model A is correct that this is a classic neurofibrillary tangle, characteristic of Alzheimer's disease neuropathology change. Model B is incorrect and hallucinates a "ballon" shaped cytoplasmic inclusion. This is a classic basophilic, flame-shaped inclusion characteristic of neurofibrillary tangle* (Supp. Figure 4). The rest consisted of preferences over Presentation and Clarity (e.g., *"Model A breaks out the information more clearly and uses better formatting.")* and Use of References and Up-to-date Guidelines (e.g., *"Model B seems to have pulled up a reference which does not exist."*). Interestingly, although we expected the presence of accurate references and citations to significantly influence clinician preference, we found that this only explained about 10% of reasons, in addition to the style analysis that showed that the use of citations were not a statistically significant predictor of model rankings. More examples are included in Figure 6.

The clinician-provided reasons also vary significantly within query categories (Figure 4): for instance, in Clinical Documentation and Practice Information, Depth and Detail is the single largest cited reason for model preference (75%), while Presentation and Clarity accounts for 60% of the cited reasons for queries related to Patient Communication and Education.

*Internal medicine, neurology, and infectious disease are the most common query subspecialties*

When categorizing queries by medical subspecialty (Figure 3, Supp. Figure 5), we find that after General Internal Medicine (35%), the top four specialties include Neurology and Neuropsychiatric Disorders (10%), Infectious Diseases (7%), Cardiology (7%), and Imaging-based Medicine (5%). Over a third of queries fall into other subspecialties, including Surgery, Public Health/Epidemiology, and Pediatrics. Google Gemini 2.0 Flash Thinking is the top-performing model across most subspecialties, while other models lead in specific domains, such as GPT-4o in imaging-based medicine and Gemini 2.5 Pro in neurology and neuropsychiatric disorders (Supp. Table 3).

**Discussion**

Early results from the MedArena platform underscore a divergence between real clinical use and what current benchmarks measure. The diversity of user preferences emphasizes that model evaluation should go beyond correctness. Clinicians frequently cited qualities like "depth and detail" and "clarity of presentation" as driving their preferences — features that are essential for trust and utility but are not captured by current automated metrics. Interestingly, stylistic elements such as formatting (e.g., bolding, lists) were found to significantly influence model preference, revealing that perceived usability and readability play a non-trivial role in model evaluation. This introduces a key challenge: how to disentangle true model reasoning quality from superficial presentation enhancements in future evaluation frameworks. These findings also highlight the value of using paired comparisons and ranking models in head-to-head scenarios, rather than relying on static accuracy scores. The Bradley-Terry model enables a more nuanced analysis by accounting for confounding factors such as response length and formatting.

Compared to the popular LM Arena, where most users are laypersons and typically spend only a few seconds on their preferences, Clinicians on MedArena spend an average of 2.5 minutes (median of 55 seconds) and 90th percentile spends over 5.5 minutes evaluating a pair of responses. This reflects the complexity and nuance of evaluating medical queries, as well as the greater attention to response differences by expert clinicians.

MedArena has several current limitations. First, due to the platform being restricted to licensed medical experts and its relatively new release, individual model matchups have a smaller sample size and thus may have less statistical significance between model ratings. Second, MedArena currently offers only 12 commercial LLMs and excludes institution- or medical-specific adapted LLMs. Starting April 2, 2025, two additional models were added to the MedArena platform: OpenAI's GPT-4.5 and Google's Gemini 2.5 Pro. While these two models are among the top performing on the leaderboard, they are not significantly better than their earlier counterparts (i.e., GPT-4o and Gemini 2.0 Flash Thinking) and have fewer preferences over their responses. Given the dynamic and rapid release of new frontier models, our initial evaluation set of models do not include sufficient sample size to report results from some of the most recent LLMs. However, the flexibility of the MedArena platform allows for continuously updating it with frontier LLMs going forward. Third, MedArena excludes the use of PHI in queries, limiting our ability to test models on complex, data-rich cases containing complete patient information. Finally, MedArena captures subjective pairwise preferences rather than ground-truth clinical correctness, so incorrect but persuasive answers could be over-preferred.

MedArena provides a scalable, clinician-centric framework for evaluating LLMs in medicine. As these tools increasingly enter clinical workflows, we hope that platforms like MedArena will improve the way that clinical medicine is evaluated in a manner that reflects the nuanced, contextual nature of real-world medical practice.

# Figures and Tables

**Figure 1**: An example query on the MedArena platform. Upon submitting a medical query, users are shown anonymized, side-by-side responses from two randomly selected models. Users then select their preferred response ("Model A," "Model B," "Tie," or "Neither") and may optionally provide a free-text explanation for their choice. MedArena supports both single- and multi-turn queries, including image-based queries. The platform includes twelve top-performing, commercially available LLMs from major providers.

| Model | BT Rating | BT CI (95%) | Win Rate | P-value vs Next |
|---|---|---|---|---|
| google/gemini-2.0-flash-thinking | 1128 | -35/+45 | 0.581 | 0.149 |
| openai/gpt-4o-2024-11-20 | 1103 | -29/+35 | 0.531 | 0.047* |
| perplexity/llama-3.1-sonar-large-128k-online | 1021 | -38/+43 | 0.412 | 0.136 |
| google/gemini-2.0-flash | 1007 | -48/+49 | 0.414 | 0.355 |
| openai/o3-mini | 992 | -37/+48 | 0.353 | 0.657 |
| meta-llama/llama-3.3-70b-instruct | 989 | -50/+52 | 0.382 | 0.131 |
| openai/o1 | 981 | -47/+52 | 0.379 | 0.93 |
| meta-llama/llama-3.2-90b-vision-instruct | 952 | -53/+56 | 0.333 | 0.211 |
| anthropic/claude-3.5-sonnet:beta | 865 | -53/+56 | 0.233 | 1 |
| google/gemini-flash-1.5 | 792 | -251/+151 | 0.2 | None |

**Table 1**: The MedArena leaderboard from November 1, 2025. From the clinician preferences collected, we calculate a ranking of ten models based on the Bradley-Terry ratings (two models with insufficient sample size were excluded). Win rates, bootstrapped 95% confidence intervals, and pairwise p-values are also calculated to assess statistical significance in model comparisons.

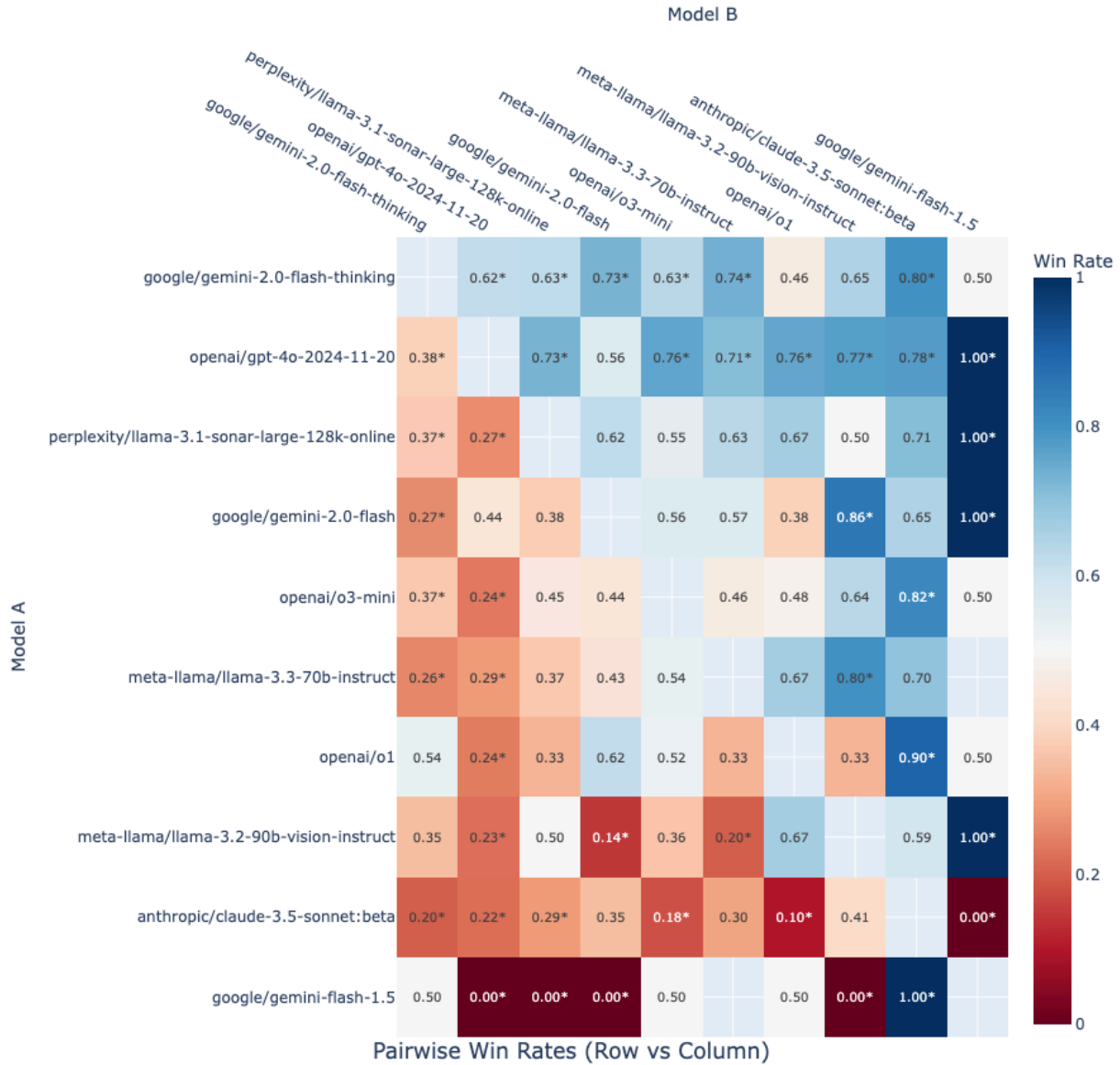

**Figure 2**: Pairwise clinician preference win rates across medical LLMs in MedArena. Heatmap showing head-to-head win rates between pairs of models, where each cell represents the probability that the row model was preferred over the column model by clinicians for a given medical query. Values greater than 0.5 indicate a preference for the row model, with darker blue denoting higher win rates and darker red denoting lower win rates; asterisks indicate statistically significant differences.

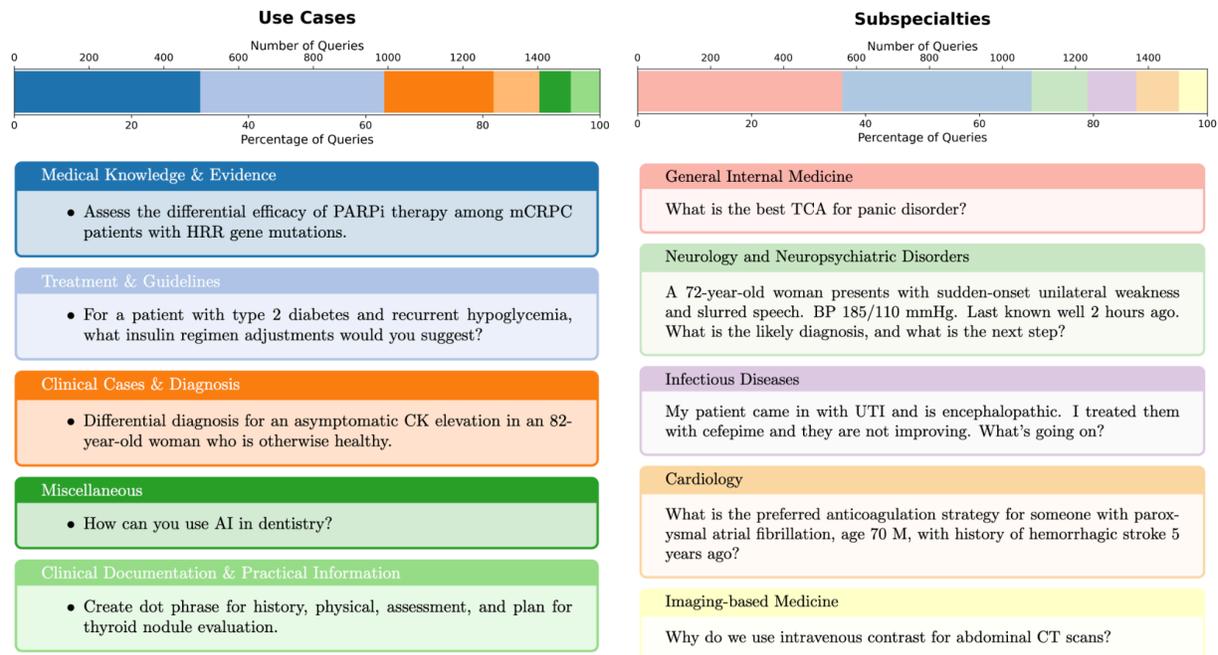

**Figure 3**: Distribution of real-world clinician queries by use case and medical subspecialty in MedArena. Left: Breakdown of clinician-submitted queries by use case. While a plurality of queries correspond to traditional medical knowledge and evidence retrieval, the majority involve treatment and guideline interpretation, clinical case reasoning, clinical documentation, patient communication, and other practical tasks. Representative examples highlight the open-ended, context-rich nature of queries encountered in routine clinical practice. Right: Distribution of queries across medical subspecialties, with General Internal Medicine comprising the largest share, followed by Neurology and Neuropsychiatric Disorders, Infectious Diseases, Cardiology, Imaging-based Medicine, and a diverse set of other specialties. Together, these distributions emphasize the breadth of clinical workflows captured by MedArena.

## Preference Reasons

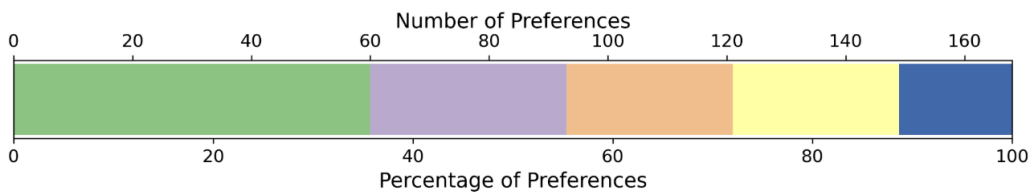

**Depth and Detail**

Model A has more detailed instructions including the nuance of if they have psoriasis

**Accuracy and Clinical Validity**

B completely missed the point.

**Miscellaneous**

Both pretty good.

**Presentation and Clarity**

Model A has a more systematic approach to treatment.

**Use of References and Up-to-Date Guidelines**

there was a recent guideline update in 2024 which suggests a threshold of 18 mmol/L to start bicarbonate supplementation

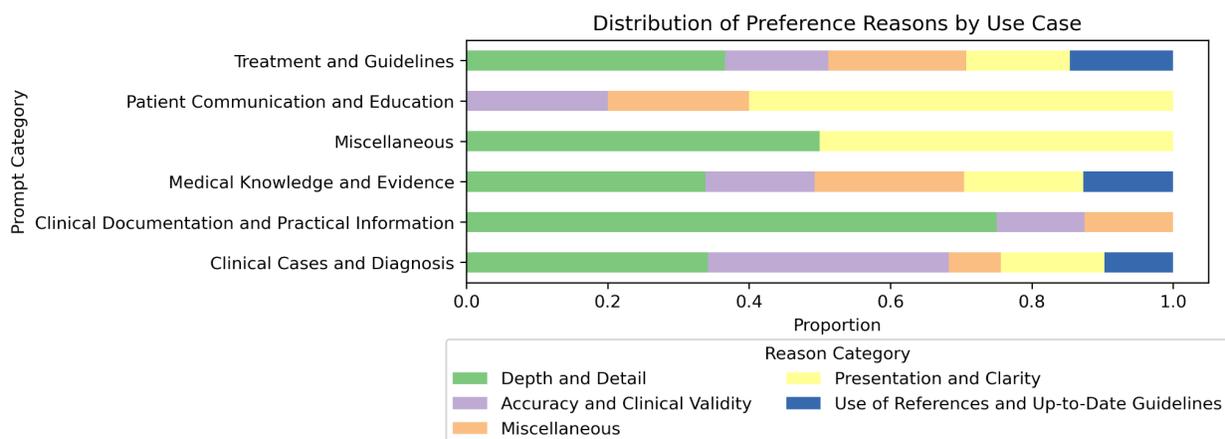

**Figure 4**: Top: Distribution of free-text preference reasons provided by clinicians, grouped into five categories: Depth and Detail, Accuracy and Clinical Validity, Presentation and Clarity, Use of

References and Up-to-Date Guidelines, and Miscellaneous. Depth and Detail is the most frequently cited reason, followed by Accuracy and Clinical Validity, highlighting clinician emphasis on thorough explanations and completeness in model responses. Representative anonymized examples illustrate the types of reasoning clinicians provided.

Bottom: Breakdown of preference reasons stratified by query use case reveals substantial heterogeneity across clinical tasks. Depth and Detail dominate in Clinical Documentation and Practical Information, while Presentation and Clarity is more prominent in Patient Communication and Education, demonstrating that clinician evaluation criteria vary with clinical context.

# References

1. Liu F, Zhou H, Gu B, et al. Application of large language models in medicine. Nat Rev Bioeng 2025;1–20.

2. Jin Q, Chen F, Zhou Y, et al. Hidden flaws behind expert-level accuracy of multimodal GPT-4 vision in medicine. NPJ Digit Med 2024;7(1):190.

3. Singhal K, Azizi S, Tu T, et al. Large language models encode clinical knowledge. Nature 2023;620(7972):172–80.

4. Bedi S, Cui H, Fuentes M, et al. MedHELM: Holistic evaluation of large language models for medical tasks [Internet]. arXiv [cs.CL]. 2025;Available from: http://arxiv.org/abs/2505.23802

5. Wu C, Qiu P, Liu J, et al. Towards evaluating and building versatile large language models for medicine. NPJ Digit Med 2025;8(1):58.

6. Chiang W-L, Zheng L, Sheng Y, et al. Chatbot Arena: An open platform for evaluating LLMs by human preference [Internet]. arXiv [cs.AI]. 2024 [cited 2025 Apr 10];Available from: http://arxiv.org/abs/2403.04132

7. Singh S, Nan Y, Wang A, et al. The Leaderboard Illusion [Internet]. arXiv [cs.AI]. 2025;Available from: http://arxiv.org/abs/2504.20879

8. Gemini Team. Gemini: A Family of Highly Capable Multimodal Models [Internet]. arXiv [cs.CL]. 2023;Available from: http://arxiv.org/abs/2312.11805

9. Achiam J, Adler S, Agarwal S, et al. Gpt-4 technical report. arXiv preprint arXiv:2303 08774 2023;

10. Touvron H, Martin L, Stone K, et al. Llama 2: Open foundation and fine-tuned chat models. arXiv preprint arXiv:2307 09288 2023;

11. Boubdir M, Kim E, Ermiş B, Hooker S, Fadaee M. Elo uncovered: Robustness and best practices in language model evaluation. IEEE Game Entertain Media Conf 2023;abs/2311.17295:106135–61.

12. Hunter DR. MM algorithms for generalized Bradley-Terry models. Ann Stat 2004;32(1):384–406.

# Supplemental Tables and Figures

| Feature | Coefficient | CI (95%) | P-value |
|---|---|---|---|
| Bold Text | 0.631 | (0.297, 1.008) | 0.000* |
| Lists | -0.399 | (-0.751, -0.053) | 0.022* |
| Token Length | 0.118 | (-0.073, 0.322) | 0.262 |
| Citations | 0.113 | (-0.096, 0.316) | 0.292 |
| Headers | 0.085 | (-0.057, 0.219) | 0.23 |

**Supplemental Table 1:** Style feature importance in clinician preference modeling. Coefficients from the extended Bradley–Terry model quantifying the association between response-level stylistic features and clinician preferences, with 95% confidence intervals and two-sided p-values. Bold text usage is positively associated with clinician preference, while the use of lists shows a modest negative association (asterisk indicates statistical significance). In contrast, token length, citation presence, and header usage are not statistically significant predictors. These results suggest that certain formatting choices influence perceived response quality, but that overall model rankings are not driven by superficial stylistic factors alone.

| Model | Elo Rating | Elo CI (95%) | BT Rating | BT CI (95%) | Style BT Rating | Style BT CI (95%) |
|---|---|---|---|---|---|---|
| openai/gpt-4o-2024-11-20 | 1072 | -33/+33 | 1119 | -38/+30 | 1110 | -35/+40 |
| googlegemini-2.0-flash-thinking | 1102 | -27/+34 | 1139 | -37/+48 | 1093 | -49/+53 |
| perplexity/llama-3.1-sonar-large-128k-online | 1009 | -28/+26 | 1041 | -42/+38 | 1051 | -44/+43 |
| google/gemini-2.0-flash | 1008 | -29/+29 | 1031 | -45/+65 | 1023 | -52/+54 |
| openai/o3-mini | 981 | -34/+27 | 1011 | -32/+31 | 1022 | -36/+45 |
| meta-llama/llama-3.3-70b-instruct | 986 | -31/+23 | 1004 | -45/+54 | 995 | -53/+54 |
| openai/o1 | 977 | -33/+36 | 987 | -55/+72 | 994 | -51/+59 |
| meta-llama/llama-3.2-90b-vision-instruct | 982 | -24/+31 | 976 | -48/+55 | 964 | -56/+56 |
| anthropic/claude-3.5-sonnet:beta | 902 | -29/+29 | 882 | -48/+61 | 932 | -63/+65 |
| google/gemini-flash-1.5 | 979 | -18/+15 | 810 | -234/+141 | 823 | -223/+140 |

**Supplemental Table 2:** Full model rankings under Elo, Bradley–Terry, and style-controlled Bradley–Terry analyses. Comparison of model performance across three ranking methods:

standard Elo ratings, Bradley–Terry (BT) ratings, and BT ratings adjusted for style-related covariates (response length, headers, lists, bold text, and citations). For each model, we report point estimates and 95% confidence intervals. Rankings are largely consistent across methods, indicating robustness of clinician preferences to the choice of ranking algorithm and to controlling for superficial stylistic factors. Minor shifts among top-ranked models after style control suggest that overall leaderboard ordering primarily reflects perceived clinical utility rather than formatting or presentation differences.

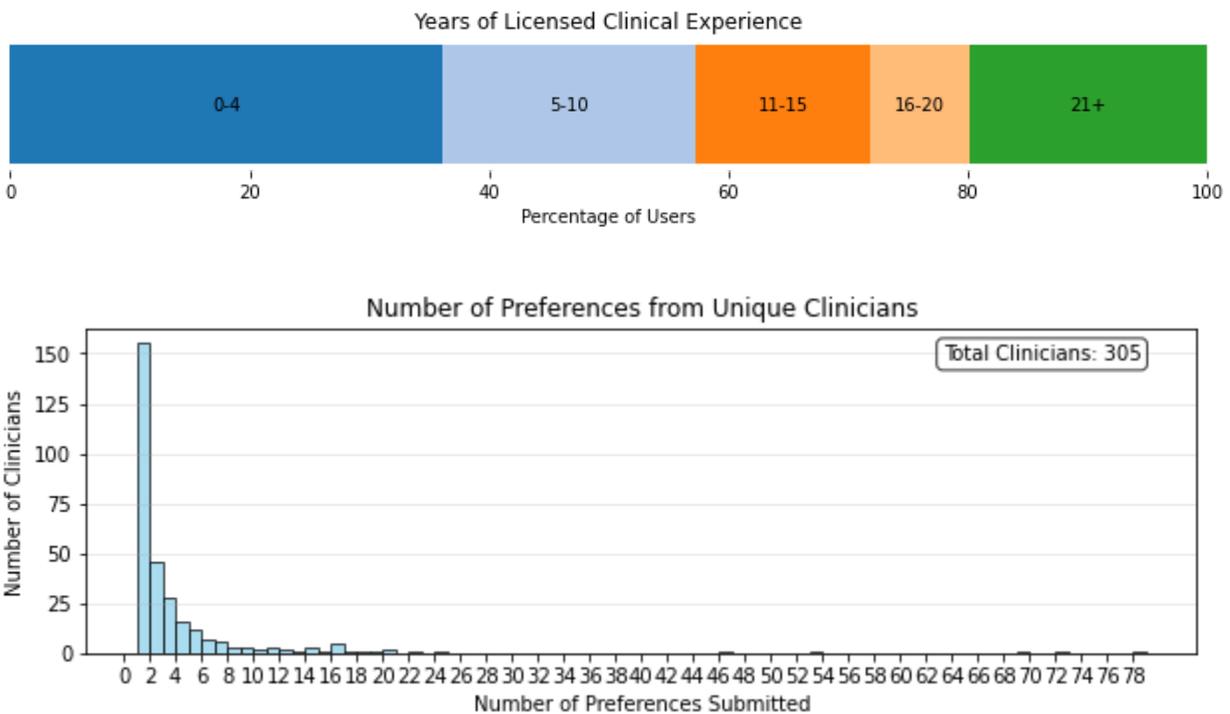

**Supplemental Figure 1**: Top: Distribution of years of licensed clinical experience among MedArena users, demonstrating participation from clinicians across career stages, from early-career practitioners to those with more than 20 years of experience.

Bottom: Histogram of the number of preferences submitted per unique clinician (N = 305), showing a long-tailed distribution in which most clinicians contributed a small number of evaluations while a subset of highly engaged users provided many preferences. Together, these results indicate that MedArena captures preferences from a diverse and experienced clinician population while enabling repeated, in-depth engagement.

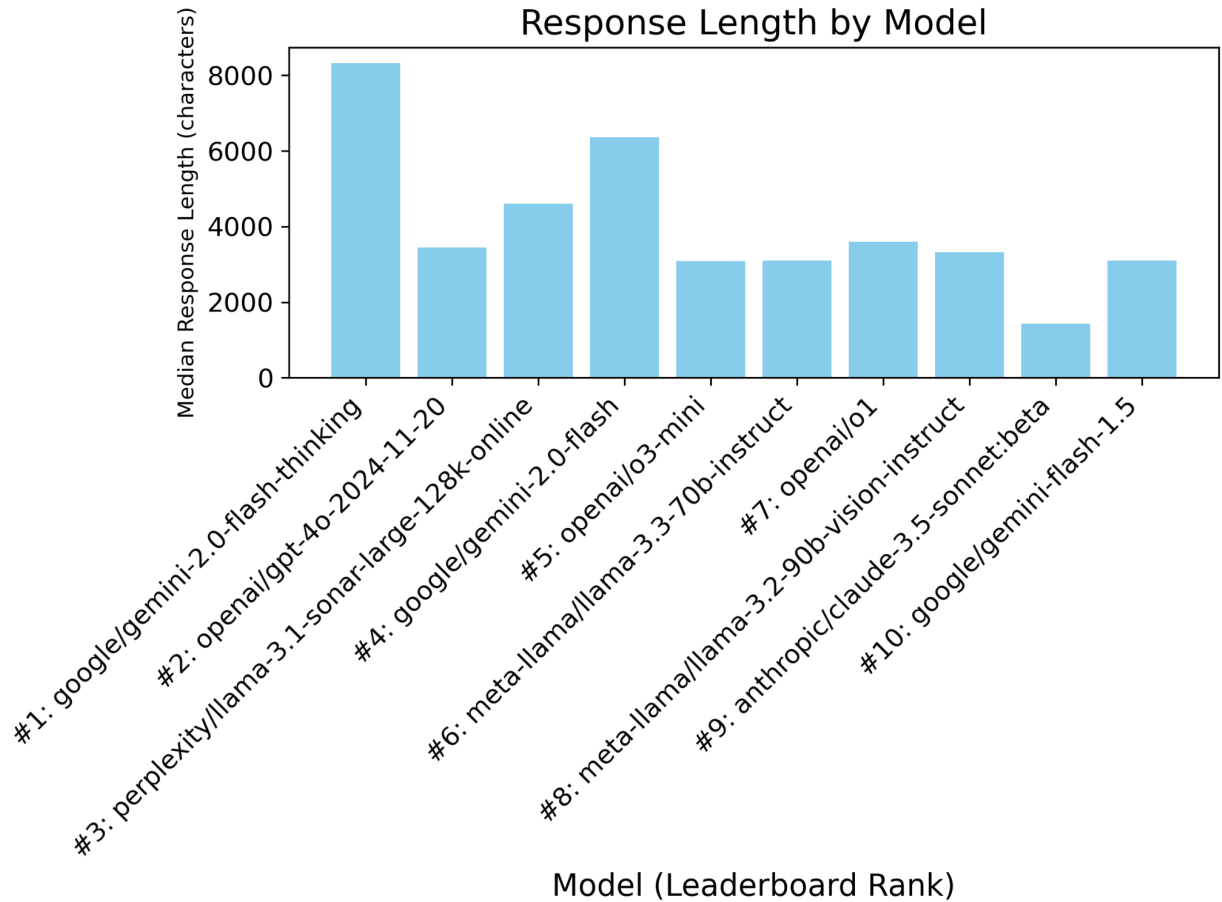

**Supplemental Figure 2**: Median response length (in characters) for each model, ordered by overall leaderboard rank. Substantial variation exists in response verbosity across models, with some higher-ranked models producing longer outputs on average. While preferred responses tend to be slightly longer, response length alone does not explain clinician preferences or overall rankings when controlling for other stylistic and content-related factors.

| Subspecialty | Top Model | Wins | Total | Win Rate |
|---|---|---|---|---|
| Cardiology | google/gemini-2.0-flash-thinking | 20 | 27 | 0.741 |
| Imaging-based Medicine | openai/gpt-4o-2024-11-20 | 16 | 24 | 0.667 |
| Infectious Diseases | google/gemini-2.0-flash-thinking | 20 | 30 | 0.667 |
| Other Specialties | google/gemini-2.0-flash-thinking | 74 | 111 | 0.667 |
| General Internal Medicine | google/gemini-2.0-flash-thinking | 85 | 132 | 0.644 |
| Neurology and Neuropsychiatric Disorders | google/gemini-2.5-pro | 8 | 13 | 0.615 |

**Supplemental Table 3**: Top-performing model by medical subspecialty. For each subspecialty, the model with the highest observed win rate is reported along with the number of wins, total comparisons, and resulting win rate.

### Clinical Cases & Diagnosis Examples

- Differential diagnosis for an asymptomatic CK elevation in an 82-year-old woman who is otherwise healthy.
- Your patient needs to start anticoagulation after a stroke 3 days ago. What other information is needed to decide when to start the anticoagulation?
- A 22-year-old college student with a 10-day history of dry cough, low-grade fever, fatigue, sore throat, headache, mild shortness of breath, scattered crackles, and wheezes; patchy interstitial infiltrate on X-ray.
- A 62-year-old woman presents with right-sided facial droop, slurred speech, and mild arm weakness (NIHSS 4). Provide diagnoses, immediate management steps, optimal imaging, and secondary prevention strategies.
- Hematopathologist assessment of myeloid neoplasm with detailed blood and marrow findings; provide top 5 differential diagnoses using ICC-2022 and WHO-HEME5 classification.

### Treatment & Guidelines Examples

- For a patient with type 2 diabetes and recurrent hypoglycemia, what insulin regimen adjustments would you suggest?

- Patient on ceftriaxone and doxycycline for CAP; suggest oral antibiotics for discharge.

- Duration of dual antiplatelet therapy after left coronary artery stenting?

- For a patient with type 2 diabetes and recurrent hypoglycemia, what insulin regimen adjustments would you suggest?

- What is the most updated guideline for treating insomnia?

## Medical Knowledge & Evidence Examples

- Assess the differential efficacy of PARPi therapy among mCRPC patients with HRR gene mutations.
- Latest scientific evidence on caloric restriction and healthy ageing.
- Construct a 15-minute journal club presentation format for novel treatments in status epilepticus.
- IBD epidemiology in the UK.
- Pathophysiology of autoinflammatory disease with a conceptual framework for physicians.

## Patient Communication & Education Examples

- Draft a sympathetic patient message regarding migraines and scheduling neurology appointment.
- Explain risks of tenecteplase for acute ischemic stroke to a patient in simple terms.
- Explain levothyroxine to a patient.
- Simple explanation to family about ALS prognosis.
- Explain to a patient with seizures the importance of taking Keppra.

> **Clinical Documentation & Practical Information Examples**
>
> - Create dot phrase for history, physical, assessment, and plan for thyroid nodule evaluation.
>
> - Dot phrase for management of heart failure exacerbation including assessment and plan.
>
> - Appeal letter to insurance company for denial of empagliflozin for worsening heart failure, including citations.
>
> - How to maximize billing in outpatient clinic.
>
> - Structured template (dot phrase) for H&P, assessment, and plan for ANCA vasculitis.

**Supplemental Figure 3**: Five anonymized queries are sampled and shown from each use case category.

## Accuracy and Clinical Validity

- Model A picked up the right diagnosis.

- Model A is correct that this is a classic neurofibrillary tangle, characteristic of Alzheimer's disease neuropathology change. Model B is incorrect and hallucinates a *"ballon" shaped cytoplasmic inclusion*. This is a classic basophilic, flame-shaped inclusion characteristic of neurofibrillary tangle.

- Both are wrong. Model A is wrong and hallucinating neuronal intranuclear inclusions and concluding HSV. These are basophilic neuronal cytoplasmic inclusions characteristic of Pick bodies in Pick's disease (frontotemporal lobar degeneration). Model B is way off and presumes this is liver tissue when it is brain tissue. It also hallucinates ballooning hepatocytes, which is completely wrong.

- Model B - it appears model A was hallucinating about NEJM paper.

- Both are wrong. This is a pyramidal neuron in the hippocampus with granulovacuolar degeneration in the cytoplasm. There is no intranuclear inclusion. The image does not show *"Negri bodies"*.

## Depth and Detail

- Model A breaks down the causes better.

- Model A has more detailed information.

- Model B has more detailed information in terms of predictors.

- Model A provides more context and more specific information.

- Model B is a bit more detailed.

### Presentation and Clarity

- Model A breaks out the information more clearly and uses better formatting.
- Formatting is much better in model A and has more information instead of just a list.
- B is a bit unwieldy - I prefer a brief answer *"cardiac causes (unlikely if asymptomatic)"*.
- Model A is presented in a more patient-friendly manner.
- Essentially a tie, but clearer formatting.

### Use of References and Up-to-date Guidelines

- Model A has references which can be useful if interested in additional information.
- Model B seems to have pulled up a reference which does not exist.
- There was a recent guideline update in 2024 which suggests a threshold of 18 mmol/L to start bicarbonate supplementation.
- Would prefer B, but sources not given.

**Supplemental Figure 4:** Five anonymized queries are sampled and shown from each reason preference type.

## General Internal Medicine

- What is the differential diagnosis for someone with a CK of 6000?
- how common is selenium deficiency
- what are the major presentations for hammer esophagus
- Teach me about POTS syndrome?
- How is moderate asthma treated?

## Other Specialties

- Best treatment option for tennis elbow from: Prp, radial shockwave, class 4 laser, cortisone, red light
- What is an uncommon type of stroke which can be related to ECMO?
- What is the intubating dose for propofol?
- I am a physician. Give me a list of 15 ways I could use AI/LLM to improve my workflow.
- Fgfr3 inactivation disease

## Neurology and Neuropsychiatric Disorders

- Can metformin help to treat alzheimer?
- bipolar 8
- Design a brief educational session on neuroanatomical correlations in brainstem lesions using cross-sectional imaging to explain clinical syndromes.
- What is the best antipsychotic medication for an admitted patient with delirium and long QTc?
- How do the 2021 CIDP diagnosis guidelines differ from the 2010 ones?

### Infectious Diseases

- Histoplasmosis punch-out   white dot syndrome wF    ?
- My patient presented with recurrent bilateral cellulitis. What are common causes of cellulitis
- which azole is less likely to cause hepatotoxicity?
- How do we treat NDM Pseudomonas malignant otitis externa
- How can I diagnose an acute EBV infection?

### Cardiology

- When is ICD placement indicated for a patient with hypertrophic cardiomyopathy
- What is the mortality rate of out of hospital cardiac arrest?
- are there any trials in asymptomatic aortic regurgitation
- Your patient received tenecteplase. What is an appropriate level of monitoring after this medication is given?
- what is the biggest modifyabil risk factor for storke

### Imaging-based Medicine

- ectatic sigmoid sinus radiology
- Imagine bilateral vertebral arteries end in PICA. How does the brainstem get blood supply?
- What is the fork sign in neurorradiology
- Shearwave elastography e median 7.95, c s median 1.61 m/sec F1 metavir score interpretation
- Over what period of time do subdural hematomas resolve on CT scans?

**Supplemental Figure 5**: Five anonymized queries are sampled and shown from each subspecialty category.

===============================================================================

Subspecialty Categorization - Full LLM Prompt

===============================================================================

System Message:

Your task is to categorize a medical question into one of these subspecialties:

1. General Internal Medicine

Description:

Covers diagnosis and management of adult medical conditions, including subspecialties such as cardiology, endocrinology, nephrology, rheumatology, gastroenterology, and pulmonology.

_____

2. Imaging-based Medicine

Description:

Encompasses interpretation and application of radiologic imaging (e.g., CT, MRI, X-ray, ultrasound) and pathological analysis (e.g., histopathology, hematopathology) in clinical decision-making.

_____

3. Cardiology

Description:

Focuses on the diagnosis, treatment, and prevention of diseases and conditions of the heart and blood vessels, including arrhythmias, heart failure, coronary artery disease, and hypertension.

_____

4. Neurology and Neuropsychiatric Disorders

Description:

Addresses disorders of the central and peripheral nervous systems, including stroke, epilepsy, neurodegenerative diseases, and neurobehavioral syndromes.

_____

5. Infectious Diseases

Description:

Covers diagnosis, treatment, and prevention of infections caused by bacteria, viruses, fungi, and parasites, as well as issues related to antibiotics, vaccines, and emerging pathogens.

-----

6. Other Specialties

Description:

Includes interdisciplinary, procedural, administrative, or uncategorized topics such as clinical documentation, CPT coding, medical devices, AI in medicine, uncommon presentations, and specialties not covered above.

Respond with ONLY the number (1-6) that best matches the subspecialty.

================================================================

Prompt Category Categorization - Full LLM Prompt

================================================================

System Message:

Your task is to categorize a medical question into one of these categories:

1. Medical Knowledge and Evidence

2. Treatment and Guidelines

3. Clinical Cases and Diagnosis

4. Patient Communication and Education

5. Clinical Documentation and Practical Information

6. Miscellaneous

Respond with ONLY the number (1-6) that best matches the category.

================================================================

Reason Category Categorization - Full LLM Prompt

================================================================

System Message:

Your task is to categorize a reason for preferring one medical response over another into one of these categories:

1. Accuracy and Clinical Validity: The preferred response is more accurate and clinically valid (e.g., "Model A is right")

2. Depth and Detail: The preferred response provides more depth and detail.

3. Presentation and Clarity: The preferred response is better presented and easier to understand.

4. Use of References and Up-to-Date Guidelines: The preferred response uses references and up-to-date guidelines.

5. Miscellaneous: The preferred response is not categorized in the above categories.

Respond with ONLY the number (1-5) that best matches the category.

**Supplemental Figure 6**: Full prompts used for generating categorizations for subspecialties, use cases, and preference reasons.